\documentclass[hidelinks,twoside]{article}
\usepackage[accepted]{aistats2018}
\usepackage{graphicx}
\usepackage{subfigure} 
\usepackage{amsmath}
\usepackage{breqn}
\usepackage{amsthm}
\usepackage{amsfonts}
\usepackage{xcolor}
\usepackage{balance}

\PassOptionsToPackage{sort&compress}{natbib}

\makeatletter
\let\NAT@parse\undefined
\makeatother
\usepackage[round]{natbib}
\usepackage{titlesec}

\newtheorem{definition}{Definition}[section]
\newtheorem{theorem}{Theorem}[section]

\newtheorem{proposition}[theorem]{Proposition}

\begin{document}

\twocolumn[

\aistatstitle{Fast Threshold Tests for Detecting Discrimination}

\aistatsauthor{ Emma Pierson \And Sam Corbett-Davies \And  Sharad Goel }

\aistatsaddress{ Stanford University \And  Stanford University \And Stanford University } 
]
\begin{abstract}
Threshold tests have recently been proposed as a useful method for detecting bias in lending, hiring, and policing decisions. For example, in the case of credit extensions, these tests aim to estimate the bar for granting loans to white and minority applicants, with a higher inferred threshold for minorities indicative of discrimination. This technique, however, requires fitting a complex Bayesian latent variable model for which inference is often computationally challenging. Here we develop a method for fitting threshold tests that is two orders of magnitude faster than the existing approach, reducing computation from hours to minutes. To achieve these performance gains, we introduce and analyze a flexible family of probability distributions on the interval $[0, 1]$---which we call  \emph{discriminant distributions}---that is computationally efficient to work with.
We demonstrate our technique by analyzing 2.7 million police stops of pedestrians in New York City.
\end{abstract}

\section{INTRODUCTION}

There is wide interest in detecting and quantifying bias in human decisions,
but well-known problems with traditional statistical tests of discrimination have hampered rigorous analysis.
The primary goal of such work is to determine whether decision makers apply different standards to groups defined by
race, gender, or other protected attributes---what economists call \emph{taste-based discrimination}~\citep{becker1957}.
For example, in the context of banking, 
such discrimination might mean that minorities are granted loans only when they are exceptionally creditworthy.
The key statistical challenge is that an individual's qualifications are typically only partially observed 
(e.g., researchers may not know an applicant's full credit history);
it is thus unclear whether observed disparities are attributable to discrimination or omitted variables.

To address this problem, \citet{simoiu2017} recently proposed the \emph{threshold test}, 
which considers both the decisions made (e.g., whether a loan was granted) 
and the outcomes of those decisions (e.g., whether a loan was repaid).
The test simultaneously estimates decision thresholds and risk profiles via a Bayesian latent variable model.
This approach mitigates some of the most serious statistical shortcomings of past methods.
Fitting the model, however, is computationally challenging, often requiring several hours on moderately sized datasets.
As is common in full Bayesian inference, the threshold model is typically fit with Hamiltonian Monte Carlo (HMC) sampling.
In this case, HMC involves repeatedly evaluating gradients of conditional beta distributions that are expensive to compute.

Here we introduce a family of distributions on the interval $[0,1]$---which we call \emph{discriminant distributions}---that is efficient for performing common statistical operations.
Discriminant distributions comprise a natural subset of logit-normal mixture distributions which is sufficiently expressive to approximate logit-normal and beta distributions for a wide range of parameters.
By replacing the beta distributions in the threshold test 
with discriminant distributions, we speed up inference by two orders of magnitude.

To demonstrate our method, we analyze 2.7 million police stops of pedestrians in New York City between 2008 and 2012. 
We apply the threshold test to assess possible bias in decisions to search individuals for weapons.
We also extend the threshold test to detect discrimination in the decision to stop an individual. 
For both problems (search decisions and stop decisions), our method accelerates inference by more than 75-fold. Such performance gains are consequential
in part because each new application requires running the threshold test dozens of times to conduct a battery of standard robustness checks. 
To carry out the experiments in this paper, we ran the threshold test nearly 100 times. 
That translates into about two months of continuous, serial computation under the standard fitting method; our approach required less than one day of computation. 
These performance gains also allow one to run the threshold test on very large datasets. In a national analysis of traffic stops by \citet{pierson_2017}, running the threshold test required splitting the data into state-level subsets; with our approach, one can fit a single national model on 22 million stops in 30 minutes, facilitating efficient pooling of information across states. Finally, such acceleration broadens the accessibility of the threshold test to policy analysts with limited computing resources. 
Our fast implementation of the threshold test is available online. 

\section{BACKGROUND}

\paragraph*{Traditional tests of discrimination.}
To motivate the threshold test, we review two traditional statistical tests of discrimination: the \emph{benchmark test} (or \emph{benchmarking}) and the \emph{outcome test}. 
The benchmark test analyzes the rate at which some action is taken (e.g., 
the rate at which stopped pedestrians are searched). 
Decision rates might vary across racial groups for a variety of legitimate reasons, such as race-specific differences in behavior.
One thus attempts to estimate decision rates after controlling for all 
legitimate factors. If decision rates still differ by race after such conditioning, 
the benchmark test would suggest bias. 
Though popular, this test suffers from the well-known problem of \emph{omitted variable bias}, as 
it is typically impossible for researchers to observe---and control for---all legitimate factors that might affect decisions.
For example, if evasiveness is a reliable indicator of possessing contraband, is not observed by researchers, and is differentially distributed across race groups, the benchmark test might indicate discrimination where there is none.
This concern is especially problematic for face-to-face interactions such as police stops that may rely on hard-to-quantify behavioral observations.

Addressing this shortcoming, \citet{becker1993,becker1957} proposed the outcome test, which
is based not on the rate at which decisions are made but on the \emph{hit rate} (i.e, the success rate) of those decisions.
Becker reasoned that even if 
one cannot observe the rationale for a search,
absent discrimination contraband should be found on 
searched minorities at the same rate as on searched whites.
If searches of minorities turn up weapons at lower rates than searches of
whites, it
suggests that officers are applying a double standard, 
searching minorities on the basis of less evidence.

Outcome tests, however, are also imperfect measures of discrimination \citep{ayres2002}.
Suppose that there are two, easily distinguishable types of white pedestrians: those who have a 1\% chance of carrying weapons, and those who have a 75\% chance. Similarly assume that black pedestrians have either a 1\% or 50\% chance of carrying weapons. If officers, in a race-neutral manner, search individuals who are at least 10\% likely to be carrying a weapon, then searches of whites will be successful 75\% of the time whereas searches of blacks will be successful only 50\% of the time. 
With such a race-neutral threshold, no individual is treated differently because of their race. Thus, contrary to the findings of the outcome test (which suggests discrimination against blacks due to their lower hit rate), no discrimination is present. 
This illustrates a failure of outcome tests known as the problem of infra-marginality~\citep{ayres2002,simoiu2017,anwar2006alternative,anwar2011testing,anwar2015testing,engel2008critique,arnold2017}.

\paragraph*{The threshold test.}
\label{sec:threshold}

To circumvent this problem of infra-marginality, 
the \emph{threshold test} of 
\citet{simoiu2017} attempts to directly infer race-specific search thresholds.
Though still relatively new, 
the test has already been used to analyze tens of millions of police stops across the United States~\citep{pierson_2017}.
The threshold test is based on a Bayesian latent variable model that formalizes the following stylized process of search and discovery. 
Upon stopping a pedestrian, officers observe 
the probability $p$ the individual is carrying a weapon;
this probability summarizes all the available information, such as the stopped individual's age and gender, criminal record, and behavioral indicators like nervousness and evasiveness.
Because these probabilities vary from one individual to the next,
$p$ is modeled as being drawn from a \emph{risk distribution} that depends on the stopped person's race ($r$) and the location of the stop ($d$), where
location might indicate the precinct in which the stop occurred. 
Officers deterministically conduct a search if the probability $p$ exceeds a race- and location-specific threshold ($t_{rd}$),
and if a search is conducted, a weapon is found with probability $p$. By reasoning in terms of risk distributions, one avoids the omitted variables problem by marginalizing out all unobserved variables. 
In this formulation, one need not observe the factors that led to any given decision,
and can instead infer the aggregate distribution of risk for each group.

\begin{figure}[t]
\centering
\includegraphics[width=0.5\columnwidth, trim=0 0.7cm 0 0cm]{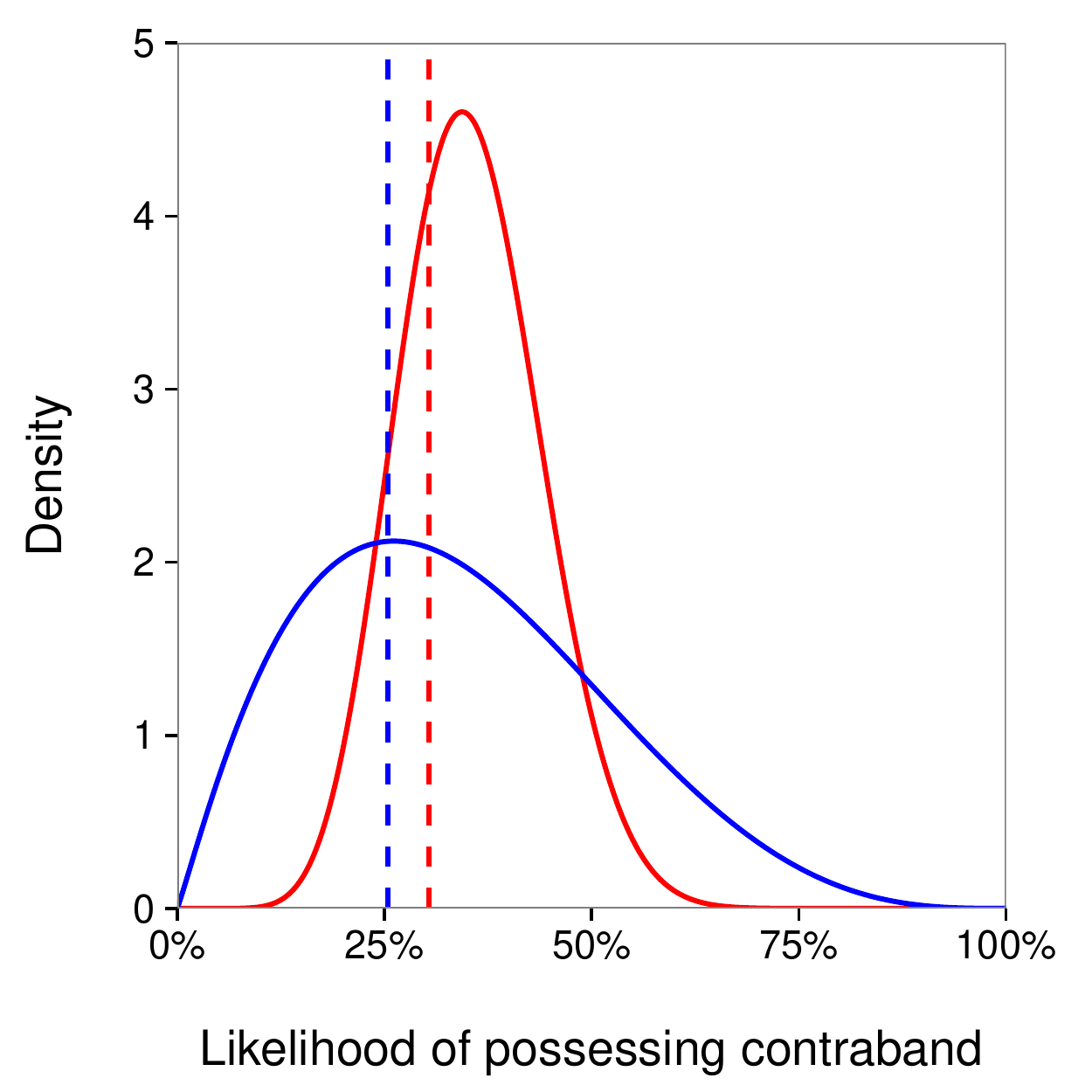}
\caption{\emph{An illustration of hypothetical risk distributions $P_{rd}$ (solid curves) and search thresholds (dashed vertical lines). The blue group is searched at a lower threshold than the red group, and so the blue group by definition faces discrimination.}}
\label{fig:threshold_test}
\vspace{-5mm}
\end{figure}

Figure \ref{fig:threshold_test} illustrates hypothetical risk distributions and thresholds for two groups in a single location. This representation visually describes the mapping from thresholds and risk distributions to search rates and hit rates (the observed data). 
Suppose $P_{rd}$ is a random variable (termed ``risk distribution'') that gives the probability of finding a weapon on a stopped pedestrian in group $r$ in location $d$.
The search rate $s_{rd}$ of group $r$ in location $d$
is $\Pr(P_{rd} > t_{rd})$, the probability a randomly selected pedestrian in that group and location exceeds the race- and location-specific search threshold;
graphically, this is the proportion of the risk distribution to the right
of the threshold.
The hit rate is the probability that a random \emph{searched} pedestrian is carrying a weapon: 
$h_{rd} = \mathbb{E}[P_{rd}\mid P_{rd}>t_{rd}]$;
in other words, the hit rate is the mean of the risk distribution 
conditional on being above the threshold. The primary goal of inference is to determine the latent thresholds $t_{rd}$. If the thresholds applied to one race group are consistently lower than the thresholds applied to another, this suggests discrimination against the group with the lower thresholds. In order to estimate the decision thresholds, 
the risk distributions must be simultaneously inferred. 
In \citet{simoiu2017}, these risk profiles take the form of beta distributions parameterized by means 
$\phi_{rd}=\textrm{logit}^{-1}(\phi_r+\phi_d)$ 
and total count parameters $\lambda_{rd} = \exp(\lambda_r+\lambda_d)$, 
where $\phi_r$, $\phi_d$, $\lambda_r$, and $\lambda_d$ are parameters that depend on the race of the stopped individuals and the location of the stops. 
Reparameterizing these risk distributions is the key to accelerating inference. Given the number of searches and hits by race and location,
we can compute the likelihood of the observed data
under any set of model parameters $\{\phi_d, \phi_r, \lambda_d, \lambda_r, t_{rd}\}$.
One can likewise compute the 
posterior distribution of the parameters given the data
and prior distributions.

\paragraph*{Inference via Hamiltonian Monte Carlo.}

Bayesian inference 
is challenging when the parameter space is high dimensional, because random walk MCMC methods fail to fully explore the complex posterior in any reasonable time. This obstacle can be addressed with Hamiltonian Monte Carlo (HMC) methods \citep{neal2011, betancourt2015hamiltonian,chen2014stochastic}, which propose new samples by numerically integrating along the gradient of the log posterior, allowing for more efficient exploration. The speed of convergence of HMC depends on three factors: (1) the gradient computation time per integration step; (2) the number of integration steps per sample; and (3) the number of effectively independent samples relative to the total number of samples (i.e., the effective sample size). The first can be improved by simplifying the analytical form of the log posterior and its derivatives. The second and third factors depend on the geometry of the posterior: a smooth posterior allows for longer paths between samples that take fewer integration steps to traverse~\citep{betancourt2017}. On all three measures, our new threshold model generally outperforms that of \citet{simoiu2017};
the improvement in gradient computation 
is particularly substantial. 

When full Bayesian inference is computationally difficult, 
it is common to consider alternatives such as variational inference~\citep{wainwright2008}.
Though fast, such alternatives have shortcomings. 
As we discuss below, variational inference produced worse fits than full Bayesian inference on our policing dataset.
Moreover, parameter estimates from variational inference varied significantly from run to run in our tests, as estimates were sensitive to initialization.
With our accelerated threshold test, one can have the best of both worlds: the statistical benefits of full Bayesian inference and the speed of fast alternatives.

\section{\mbox{DISCRIMINANT DISTRIBUTIONS}}

The computational complexity of the standard threshold test 
is in large part due to difficulties of working with beta distributions.
When $P$ has a beta distribution,
it is expensive to compute the search rate $\Pr(P > t)$, the hit rate $\mathbb{E} [P\mid P > t]$, and their associated derivatives~\citep{boik1998derivatives}. 
Here we introduce an alternative family of 
\emph{discriminant distributions} for which it is
efficient to compute these quantities.
We motivate and analyze this family in the specific context of the threshold test, but the family itself can be applied more widely.

To define discriminant distributions, assume that there are two classes (positive and negative), 
and the probability of being in the positive class is $\phi$.
For example, positive examples might correspond to individuals who are
carrying weapons, and negative examples to those who are not.
We further assume that each
class emits \emph{signals} that are normally distributed
according to $\mathrm{N}(\mu_0, \sigma_0)$ and $\mathrm{N}(\mu_1, \sigma_1)$, respectively.
Denote by $X$ the signal emitted by a random instance in the population,
and by $Y \in \{0,1\}$ its class membership.
Then, given an observed signal $x$, one can compute the
probability $g(x) = \Pr(Y = 1 \mid X =x)$ that it was emitted by a member of the positive class.
Throughout the paper, we term the domain of $g$ the \emph{signal space} and its range the \emph{probability space}.
Finally, we say the random variable $g(X)$ has a discriminant distribution
with parameters $\phi$, $\mu_0$, $\sigma_0$, $\mu_1$, and $\sigma_1$.

\begin{definition}[Discriminant distribution]
\label{def:disc}
Consider parameters $\phi \in (0,1)$, 
$\mu_0 \in \mathbb{R}$, 
$\sigma_0 \in \mathbb{R}_+$, 
$\mu_1 \in \mathbb{R}$, 
and $\sigma_1 \in \mathbb{R}_+$,
where $\mu_1 > \mu_0$.
The discriminant distribution 
$\mathrm{disc}(\phi, \mu_0, \sigma_0, \mu_1, \sigma_1)$ is defined as follows. 
Let
$$Y \sim \textrm{Bernoulli}(\phi),$$
$$X \mid Y=0\sim \mathrm{N}(\mu_0, \sigma_0),$$ 
$$X \mid Y=1\sim \mathrm{N}(\mu_1, \sigma_1).$$ 

Set $g(x) = \Pr(Y=1 \mid X=x)$. 
Then the random variable $g(X)$ is distributed as
$\mathrm{disc}(\phi, \mu_0, \sigma_0, \mu_1, \sigma_1)$.
\end{definition}

Our description above mirrors the motivation of linear discriminant analysis (LDA). 
Although it is common to consider the conditional probability of class membership $g(x)$,
it is less common to consider the \emph{distribution} of these probabilities as an alternative to beta or logit-normal distributions.
To the best of our knowledge, 
the computational properties of discriminant distributions 
have not been previously studied.

As in the case of LDA, the statistical properties of discriminant distributions are particularly nice when the underlying normal distributions have the same variance.
Proposition~\ref{prop:monotonicity} below establishes a key monotonicity property that is standard in the development of LDA;
though the statement is well-known, we include it here for completeness. Proofs of this proposition and other technical statements are in the Supplementary Information (SI). 

\begin{proposition}[Monotonicity]
\label{prop:monotonicity}
Given a discriminant distribution $\mathrm{disc}(\phi, \mu_0, \sigma_0, \mu_1, \sigma_1)$, the mapping $g$ from signal space to probability space is monotonic if and only if $\sigma_0=\sigma_1$.
\end{proposition}

We confine our attention to homoskedastic discriminant distributions so that the mapping from signal space to probability space will be monotonic. Without this property, we cannot interpret a threshold on signal space as a threshold in probability space, which is key to our analysis. Homoskedastic discriminant distributions (i.e., with $\sigma_0 = \sigma_1$) 
involve four parameters. But in fact only two parameters are required to fully describe this family of distributions.
This simplified parameterization is useful for computation.

\begin{proposition}[2-parameter representation]
\label{prop:2-parameter}
Suppose 
$\mathrm{disc}(\phi, \mu_0, \sigma, \mu_1, \sigma)$
and 
$\mathrm{disc}(\phi', \mu_0', \sigma', \mu_1', \sigma')$
are two homoskedastic discriminant distributions.
Let 
\begin{equation*}
\delta = \frac{\mu_1-\mu_0}{\sigma}
\end{equation*}
and define $\delta'$ analogously.
Then the two distributions are identical if
$\phi = \phi'$ and $\delta= \delta'$.
As a result, homoskedastic discriminant distributions 
can be parameterized by $\phi$ and $\delta$ alone.
\end{proposition}

Given Proposition~\ref{prop:2-parameter},
we henceforth write $\mathrm{disc}(\phi, \delta)$ 
to denote a homoskedastic discriminant distribution. Considering $\mathrm{disc}(\phi, \delta)$ as a distribution of calibrated predictions, the parameters have intuitive interpretations: $\phi$ is the fraction of individuals in the positive class, while $\delta$ is monotonically related to the AUC-ROC of the predictions\footnote{AUC-ROC is the probability that a random member of the positive class is assigned a higher score than a random member of the negative class. Since the positive and negative class emit signals from independent normal distributions, it's straightforward to show that the AUC-ROC equals $\Phi\left(\frac{\delta}{\sqrt{2}}\right)$ (where $\Phi(\cdot)$ is the CDF of the standard normal distribution).}.
Even though the distribution itself depends on only $\phi$ and $\delta$,
the transformation $g$ from signal space to probability space depends on the particular 4-parameter representation we use.
For simplicity, we consider the representation with
$\mu_0 = 0$ and $\sigma=1$.
This yields the simplified transformation function:
\begin{align*}
g(x)=\frac{1}{1+\frac{1-\phi}{\phi}\exp\left(-\delta x+\delta^2/2\right)}.
\end{align*}

Our primary motivation for introducing discriminant distributions
is to accelerate key computations of the (complementary) CDF and conditional means.
Letting $P = g(X)$, we are specifically interested in 
computing $\Pr(P > t)$ and
$\mathbb{E}[P\mid P > t]$. 
With (homoskedastic) discriminant distributions, 
these quantities map nicely to signal space,
where they can be computed efficiently.

Denote by $\bar \Phi(x;\mu,\sigma)$ the normal complementary CDF.
Then the complementary CDF of $P$ can be computed as follows.
\begin{align*}
\Pr(P> t) = (1-\phi)\bar \Phi(g^{-1}(t);0,1)+\phi\bar \Phi(g^{-1}(t);\delta, 1).
\end{align*}

For the conditional mean, we have
\begin{align*}
\mathbb{E}[P\mid P> t] = \frac{\phi\bar \Phi(g^{-1}(t);\delta, 1)}{(1-\phi)\bar \Phi(g^{-1}(t);0,1)+\phi\bar \Phi(g^{-1}(t);\delta, 1)}.
\end{align*}

Importantly, the CDF and conditional means for 
discriminant distributions are closely related to those for
the normal distributions, and as such are computationally efficient to work with. In particular, the gradients of these functions are relatively straightforward to evaluate.
The corresponding quantities for 
logit-normal and beta distributions involve tricky numerical approximations~\citep{frederic2008two, jones2009kumaraswamy}.

\begin{figure}[t]
\centering
\includegraphics[width=.9\columnwidth, trim=0 0.7cm 0 0cm]{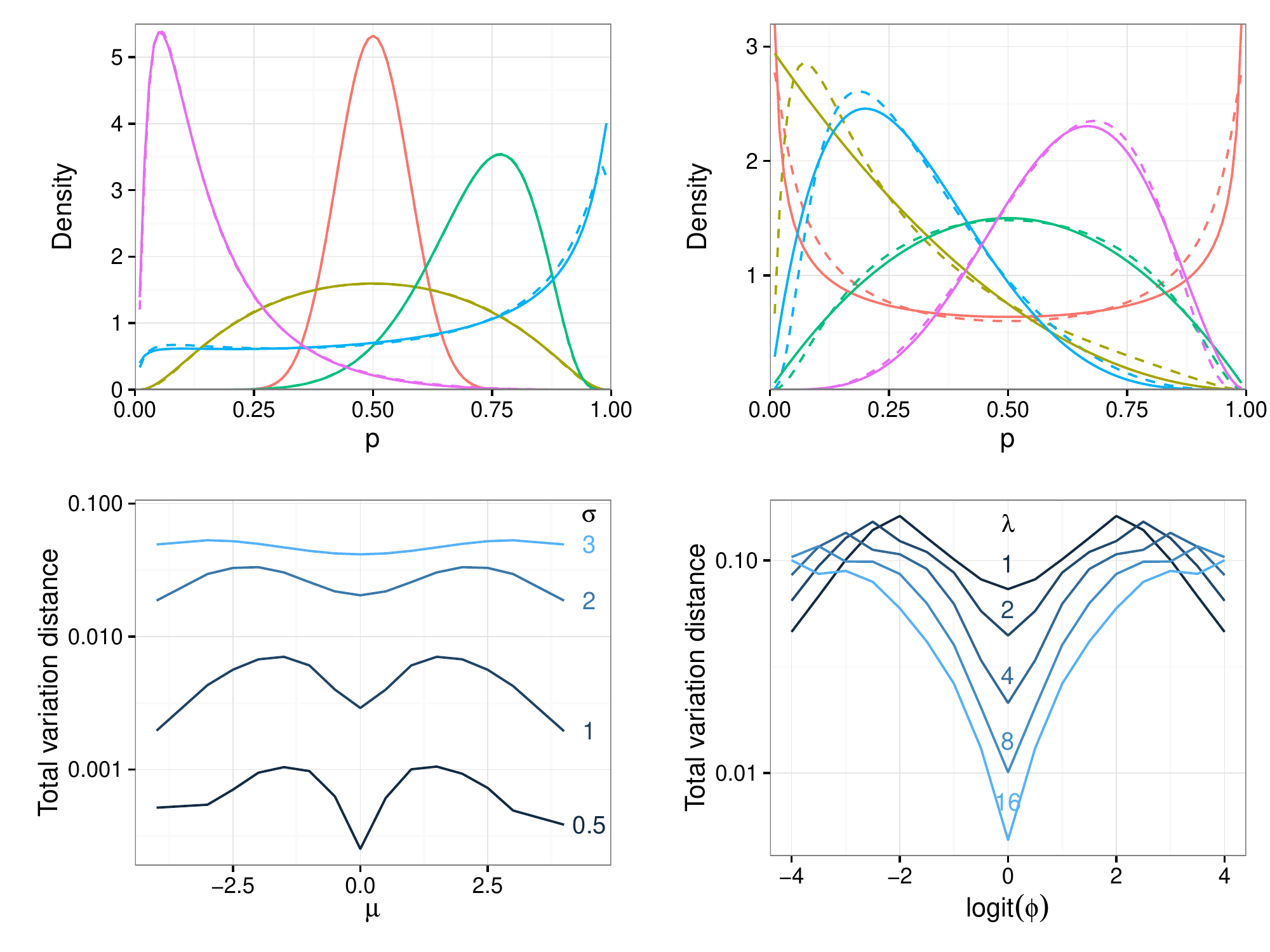}
\caption{\emph{Discriminant distributions can approximate the $\mathrm{logit}\textrm{-}\mathrm{normal}(\mu,\sigma)$ (left panels) and $\mathrm{beta}(\phi,\lambda)$ (right panels) distributions. 
In the top row we illustrate some typical distributions (solid lines) and their discriminant distribution approximations (dashed lines; some approximations are too accurate for lines to be seen). In the bottom row we explore a wide range of parameter values for distributions and report the total variation distance between the reference distribution and its discriminant distribution approximation.
}}
\label{fig:approx}
\vspace{-5mm}
\end{figure}

Finally, we show that discriminant distributions are an expressive family of distributions (Fig.~\ref{fig:approx}): they can approximate typical instantiations of the logit-normal and beta distributions. First, we select the parameters of the reference distribution: $(\mu,\sigma)$ for the logit-normal or $(\phi,\lambda)$ for the beta. Then we numerically optimize the parameters of the discriminant distribution to minimize the total variation distance between the reference distribution and the discriminant distribution. The top row of Figure ~\ref{fig:approx} shows some typical densities and their approximations. The bottom row investigates the approximation error for a wide range of parameter values. The discriminant distribution fits the logit-normal very well (distance below $0.1$ for all distributions with $\sigma\leq 3$), and the beta distribution moderately well (distance below $0.2$ for $\lambda\geq 1$). Discriminant distributions approximate logit-normal distributions particularly well because they form a subset of logit-normal mixture distributions (SI).

\section{STOP-AND-FRISK CASE STUDY}
\label{sec:application}

To demonstrate the value of discriminant distributions for 
speeding up the threshold test,
we analyze a public dataset of pedestrian stops 
conducted by New York City police officers under its
``stop-and-frisk'' practice.
Officers have legal authority to stop and briefly detain
individuals when they suspect criminal activity.
There is worry, however, that such discretionary decisions
are prone to racial bias; 
indeed the NYPD practice was recently ruled discriminatory 
in federal court and subsequently curtailed~\citep{Floyd,sklansky2017}.
Here we revisit the statistical evidence for discrimination. Our dataset contains information on 
2.7 million police stops occurring between 2008 and 2012.
Several variables are available for each stop,
including the race of the pedestrian,
the police precinct in which the stop occurred,
whether the pedestrian was ``frisked'' 
(patted-down in search of a weapon), 
and whether a weapon was found.
We analyze stops of white, black, and Hispanic pedestrians,
as there are relatively few stops of individuals of other races. We use the threshold test to analyze two decisions:
the initial stop decision, and the subsequent decision of whether or not 
to conduct a frisk.
Analyzing frisk decisions is a straightforward application of 
the threshold test: simply replacing beta distributions in the model with discriminant distributions results in more than a 100-fold speedup.
To analyze stop decisions, we extend the threshold model to the case where one does not observe negative examples (i.e., those who were not stopped) and show that discriminant distributions again produce significant speedups.

\subsection{Assessing bias in frisk decisions}

\begin{figure}[t]
\centering
\includegraphics[width=.7\columnwidth, trim=0 0.7cm 0 0cm]{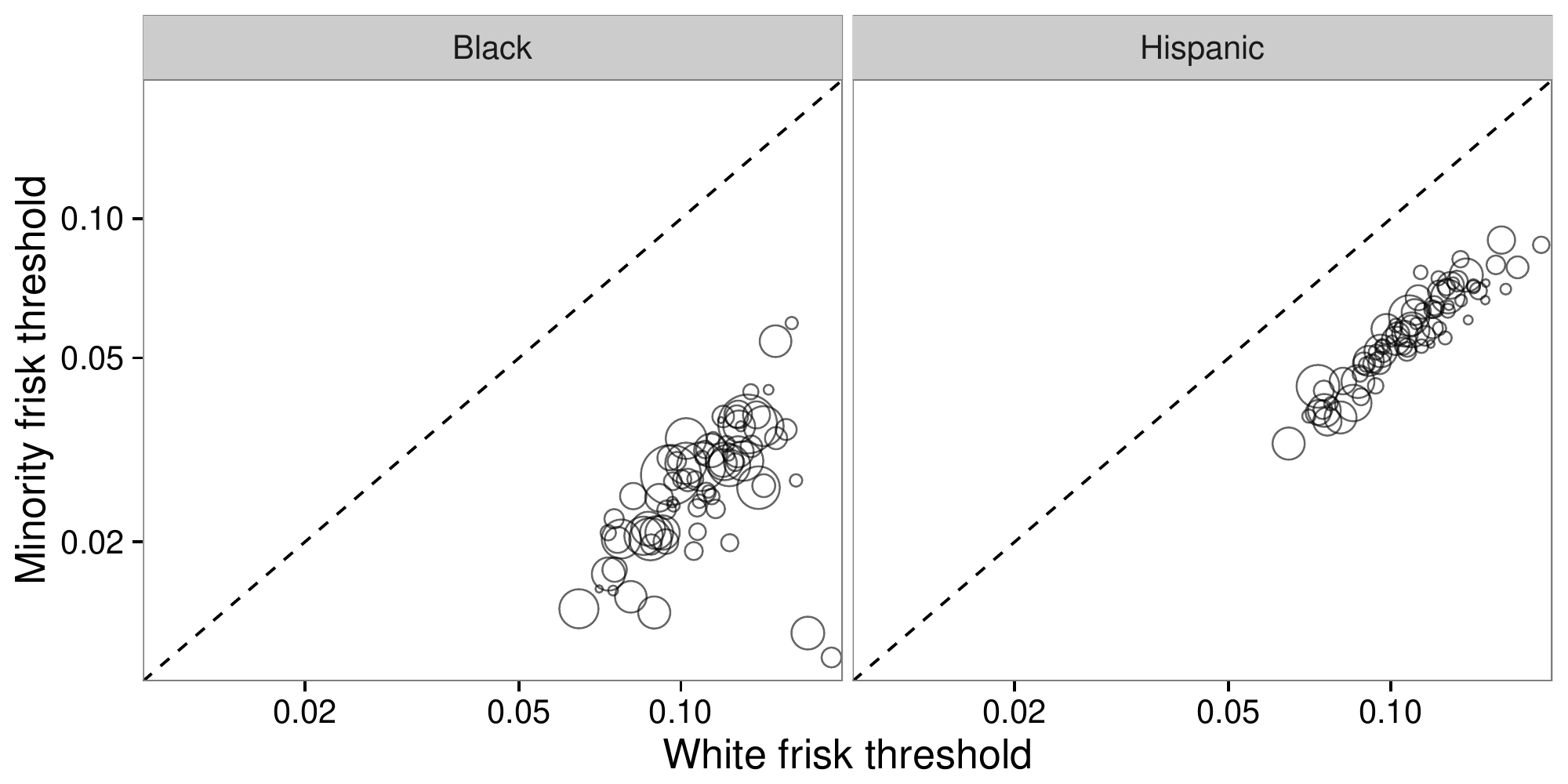}

\caption{\emph{Inferred thresholds for frisks in the stop-and-frisk data. Thresholds for white pedestrians are plotted on the horizontal axis and thresholds for minority pedestrians in the same precinct are plotted on the vertical axis. The dotted line denotes equal thresholds. The size of each circle corresponds to the number of stops of minority pedestrians. Axes are logarithmic.}}
\label{fig:stop_and_frisk_search_thresholds}
\vspace{-4mm}
\end{figure}

We fit the threshold models using Stan \citep{carpenter2016stan}, a language for full Bayesian statistical inference via HMC.
When using beta distributions in the threshold test,
it takes nearly two hours to infer the model parameters;
when we replace beta distributions with
discriminant distributions, inference completes in under one minute. Why is it that discriminant distributions result in such a dramatic increase in performance?
The compute time per effective Monte Carlo sample is the product of three terms: 
\begin{equation*}
\frac{\text{seconds}}{n_{\text{eff}}} =  \frac{\text{samples}}{n_{\text{eff}}} \cdot \frac{\text{integration steps}}{\text{sample}}\cdot
\frac{\text{seconds}}{\text{integration step}}.
\end{equation*}
All three factors are significantly reduced by using discriminant distributions (Table~\ref{tab:frisk_timing_breakdown}), with the final term providing the most significant reduction.  (The reduction in the first two terms is likely due to the geometry of the underlying parameter space and the accuracy with which we can numerically approximate gradients for the beta and discriminant distributions.) Using discriminant distributions reduces the time per effective sample by a factor of 760. In practice, one typically runs chains in parallel, 
and so total running time is determined by the last chain to terminate.
When running five chains in parallel for 5,000 iterations each,
the accelerated model is faster by a factor of 140.

The thresholds inferred using the accelerated model are extremely highly correlated with the thresholds inferred under the original model (correlation = 0.95), and indicate discrimination against black and Hispanic individuals. (In general, we would not expect the original and accelerated models to yield identical results on all datasets, since they use different probability distributions, so the fact that the thresholds are highly correlated serves as a robustness check.) Figure~\ref{fig:stop_and_frisk_search_thresholds} shows the inferred thresholds (under the accelerated model). Each point corresponds to the threshold for one precinct, with the threshold for white pedestrians on the horizontal axis and for minority pedestrians on the vertical axis. 
Within precinct, thresholds for minority pedestrians are consistently lower than thresholds for white pedestrians. 

\begin{table}[t]
\caption{\emph{Sampling times for the frisk decision model. Each row reports the improvement for the discriminant distribution model relative to the beta distribution model followed by the statistics for both models. The first row reports the seconds per effective sample, which is the product of the numbers in the second to fourth rows. The final row reports the time to fit the models used in our analysis using 5 chains run in parallel for 5,000 samples. Sampling was performed on a server with two Intel Xeon E5 processors with 16 cores each.}} 
\label{tab:frisk_timing_breakdown}
\vspace{2mm}
\begin{scriptsize}
\begin{sc}
\centerline{
\begin{tabular}{rlll}
\hline
& Speedup & Beta & Disc \\ 
  \hline
Seconds/$n_{\text{eff}}$ & \textbf{760x}  & 26  & 0.03 \\ 
  Samples/$n_{\text{eff}}$ &  \textbf{5x} & 22  &  5 \\ 
  Integration steps/sample &  \textbf{3x} & 152  & 49 \\ 
  Seconds/integration step & \textbf{53x}  & 0.008  & 0.0001  \\ 
 Seconds to fit model & \textbf{143x}   & 5,977 & 42  \\    \hline
\end{tabular}
}
\end{sc}
\end{scriptsize}
\vspace{-5mm}
\end{table}

\paragraph*{Robustness checks.}
To evaluate the robustness of our substantive finding of bias in frisk decisions, we perform a series of checks for threshold tests recommended by~\citet{simoiu2017}; we include all figures in the SI. 
We start by conducting \emph{posterior predictive checks}~\citep{gelman1996posterior}. 
We compute the model-inferred frisk and hit rates for each precinct and race group, 
and compare these to the observed rates (SI Figure 1).
The model almost perfectly fits the observed frisk rates, and fits the observed hit rates quite well: The RMSE of frisk rates is 0.05\%, and the RMSE of hit rates is 2.5\%. (RMSEs for the original beta model are comparable: the RMSE for frisk rates is 0.1\%, and the RMSE for hit rates is 2.4\%). For comparison, if the model of \citet{simoiu2017} is fit with variational inference---rather than HMC, to speed up inference---the frisk rate RMSE is 0.15\% (a 3-fold increase), and the hit rate RMSE is 2.6\% (on par with HMC). Variational inference fits the model of \citet{simoiu2017} in 44 seconds, comparable to the runtime with HMC and discriminant distributions. 

The stylized behavioral model underlying the threshold test posits a single frisk threshold for each race-precinct pair.
In reality, officers within a precinct might apply different thresholds, and even the same officer might vary the threshold from one stop to the next.
Moreover, officers only observe noisy approximations of a stopped pedestrian's likelihood of carrying a weapon;
such errors can be equivalently recast as variation in the frisk threshold applied to the true probability. 
To investigate the robustness of our results to such heterogeneity, 
we next examine the stability of our inferences on synthetic datasets derived from a generative process with varying thresholds.
We start with the model fit to the actual data.
Then, for each observed stop, we draw a signal $p$ from the inferred signal distribution for the precinct $d$ in which the stop occurred and
the race $r$ of the pedestrian.
Second, we set the stop-specific threshold to $T \sim \text{logit-normal}(\text{logit}(t_{rd}), \sigma)$, where 
$t_{rd}$ is the inferred threshold, and $\sigma$ is a parameter we set to control the degree of heterogeneity in the thresholds. This corresponds to adding normally-distributed noise to the inferred threshold on the logit scale.
Third, we assume a frisk occurs if and only if $p \geq T$, and if a frisk is conducted, we assume a weapon is found with probability $p$.
Finally, we use our modeling framework to infer new frisk thresholds $t_{rd}'$ for the synthetic dataset.

There is a steady decrease in inferred thresholds as the noise increases (SI Figure 2). Importantly, however, there is a persistent gap between whites and minorities despite this decline, indicating that the lower thresholds for minorities are robust to heterogeneity in frisk thresholds. 
$\sigma=1$ is a substantial amount of noise. Decreasing the frisk threshold of blacks by 1 on the logit scale corresponds to a 3-fold increase in the city-wide frisk rate of blacks.

In theory, the threshold test is robust to unobserved heterogeneity that affects the \emph{signal},
since we effectively marginalize over any omitted variables when estimating the signal distribution.
However, we must still worry about systematic variation in the \emph{thresholds} that is correlated with race.
For example, if officers apply a lower frisk threshold at night, and black individuals
are disproportionately likely to be stopped at night, then blacks would, on average, experience a lower frisk threshold
than whites even in the absence of discrimination.
Fortunately, as a matter of policy only a limited number of factors may legitimately affect the frisk thresholds, 
and many---but not all---of these are recorded in the data.
There are a multitude of hard-to-quantify factors (such as behavioral cues) that may affect the signal---but these should not affect the threshold.

We examine the robustness of our results to variation in thresholds across year, time-of-day, and age and gender of the stopped pedestrian.\footnote{Variation across location is explicitly captured by the model. Gender, like race, is generally not considered a valid criterion for altering the frisk threshold, though for completeness we still examine its effects on our conclusions.}
To do so, we disaggregate our primary dataset by year (and, separately, by time-of-day, by age, and by gender), and then independently run the threshold test on each component (SI Figure 3).
The inferred thresholds do indeed vary across the different subsets of the data.
However, in every case, the thresholds for frisking blacks and Hispanics are lower
than the thresholds for frisking whites, corroborating our main results.

Finally, we conduct two \emph{placebo tests}, where we rerun the threshold test with race replaced by day-of-week, and separately, 
with race replaced by month.
The hope is that the threshold test accurately captures a lack of ``discrimination'' based on these factors. The model indeed finds that the threshold for frisking individuals is 
relatively stable by day-of-week, with largely overlapping credible intervals (SI Figure 4).
We similarly find only small differences in the inferred monthly thresholds. Some variation is expected, as officers might legitimately apply slightly different frisk standards
throughout the week or year.

\subsection{Assessing bias in stop decisions}

We now extend the threshold model to test for discrimination in an officer's decision to stop a pedestrian.
In contrast to frisk decisions, we do not observe instances in which an officer decided \emph{not} to carry out a stop.
Inferring thresholds with such censored data 
is analogous to learning classifiers from only positive and unlabeled examples~\citep{elkan2008, mordelet2014bagging, du2014analysis}.
We assume officers are equally likely to encounter anyone in a precinct.
Coupled with demographic data compiled by the U.S. Census Bureau,
this assumption lets us estimate the racial distribution of individuals
encountered by officers.
Such estimates are imperfect, in part because 
residential populations differ from daytime populations~\citep{bhaduri2008population};
however, our inferences are robust to 
violations of this assumption. 

\paragraph*{Model description.}

When analyzing frisk decisions, 
the decision itself and the success of a frisk were modeled as random outcomes.
For stops, we model as random 
the race of stopped individuals and whether a stop was successful (i.e., turned up a weapon).
The likelihood of the observed outcomes can then be computed under any set of model parameters, which allows us to compute 
posterior parameter estimates. Let $S_{rd}$ denote the number of stops 
of individuals of race $r$ in precinct $d$, 
and let $H_{rd}$ denote the number of such stops that yield a weapon.
We denote by $c_{rd}$ the fraction of people in a precinct of a given race.
Letting $R_d$ denote the race of an individual randomly encountered by the police in precinct $d$, we have
\begin{equation*}
\Pr(R_d = r \mid \textrm{stopped}) \propto \Pr(\textrm{stopped} \mid R_d = r) \Pr(R_d = r).
\end{equation*}
Assuming officers are equally likely to encounter everyone in a precinct,
$\Pr(R_d = r) = c_{rd}$.
We further assume that individuals of race $r$ are stopped when their probability of carrying a weapon exceeds a race- and precinct-specific threshold $t_{rd}$; this assumption mirrors the one made for the frisk model.
Setting $\theta_{rd} = \Pr(R_d = r \mid \textrm{stopped})$,
we have
$$\theta_{rd} \propto c_{rd} \cdot \Pr(\textrm{stopped} \mid R_d = r)$$
\begin{dmath*}
\Pr(\textrm{stopped} \hiderel{\mid R_d = r}) = (1 - \phi_{rd})\bar \Phi(t_{rd}; 0, 1) \\
 + \phi_{rd}\bar \Phi(t_{rd}; \delta_{rd}, 1).
\end{dmath*}

For each precinct $d$, 
the racial composition of stops is thus distributed as a multinomial: 
\begin{equation*}
\vec{S}_{d} \sim \textrm{multinomial}\Big(\vec{\theta}_{d}, N_d\Big)
\end{equation*}
where $N_d$ denotes the total number of stops conducted in that precinct,
$\vec{\theta}_{d}$ is a vector of race-specific stop probabilities $\theta_{rd}$,
and $\vec{S}_{d}$ is the number of stops of each race group in that precinct.
We model hits as in the frisk model. 
We put normal or half-normal
priors on all the parameters: 
$\{\phi_d, \phi_r, \lambda_d, \lambda_r, t_{rd}\}$.\footnote{
Following
\citet{simoiu2017} and \citet{pierson_2017}, we 
put weakly informative priors on $\phi_r$, $\lambda_r$, and
$t_{rd}$; we put tighter priors on the location parameters
$\phi_d$ and $\lambda_d$ to restrict geographical 
heterogeneity and to accelerate convergence.
}

\paragraph*{Results.}

\begin{figure}[h]
\centering
\includegraphics[width=.7\columnwidth, trim=0 0.7cm 0 0cm]{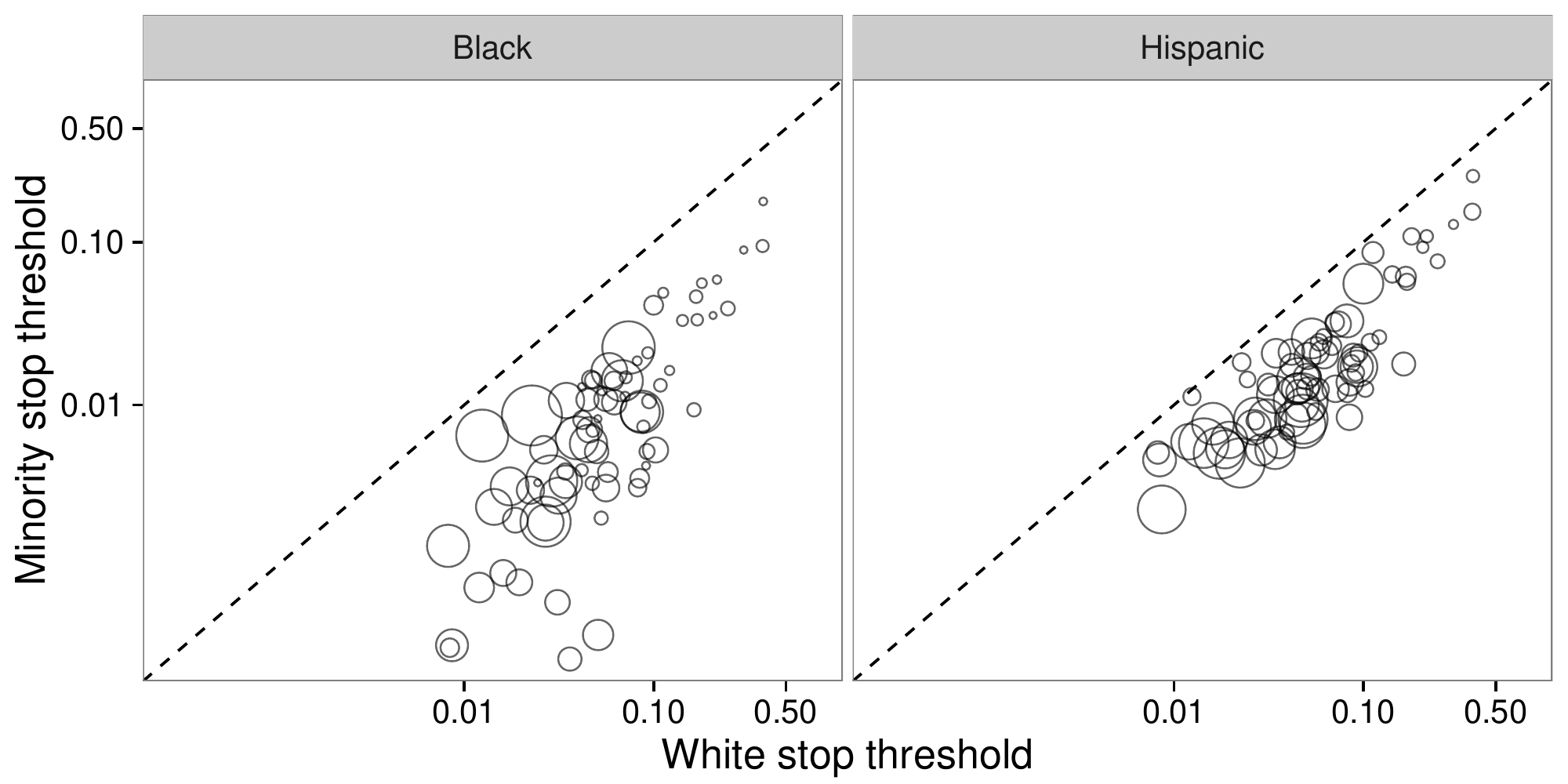}
\caption{\emph{Inferred stop thresholds in the stop-and-frisk data. The size of each circle corresponds to  the minority population in each precinct. Axes are logarithmic.}}
\label{fig:stop_and_frisk_stop_thresholds}
\vspace{-3mm}
\end{figure}

We apply the above model to the subset of approximately 723,000 stops
predicated on suspected criminal possession of a weapon,
as indicated by officers.
In these cases, the stated objective of the stop is discovery of a weapon,
and so we consider a stop successful if a weapon was discovered~\citep{goel2016precinct}.
We estimate the racial composition of precincts using data from the 2010 U.S. Census. In Table~\ref{tab:stop_timing_breakdown} we compare the time to fit our stop model
with discriminant distributions
rather than beta distributions.
The speedup from using discriminant distributions
is dramatic:
with beta distributions, the model requires more than 4 hours to fit;
with discriminant distributions it takes under 4 minutes.
The primary reason for the speedup is the reduced time per gradient evaluation, although the number of gradient evaluations is also reduced. As with frisk decisions,
we find stop thresholds for blacks and Hispanics
are consistently lower than for whites, suggestive of discrimination (Figure~\ref{fig:stop_and_frisk_stop_thresholds}).
Our results are in line with those from past statistical studies of New York City's stop-and-frisk practices based on benchmark~\citep{gelman2007analysis} and outcome~\citep{goel2016precinct} analysis.

\begin{table}[t]
\caption{\emph{Breakdown of sampling times for the stop model. Each row reports the improvement for the discriminant distribution model relative to the beta distribution model followed by the statistics for both models.}} 
\label{tab:stop_timing_breakdown}
\begin{scriptsize}
\begin{sc}
\centerline{
\begin{tabular}{rlll}
\hline
& Speedup & Beta & Disc \\ 
  \hline
Seconds/$n_{\text{eff}}$ & \textbf{84x}  & 55  & 0.7 \\ 
  Samples/$n_{\text{eff}}$ &  \textbf{1x} & 18  &  17 \\ 
  Integration steps/sample &  \textbf{2x} & 503  & 237 \\ 
  Seconds/integration step & \textbf{37x}  & 0.006  & 0.0002  \\ 
 Seconds to fit model & \textbf{77x}   & 15,943 & 208  \\    \hline
\end{tabular}
}
\end{sc}
\end{scriptsize}
\vspace{-2mm}
\end{table}

\paragraph*{Robustness checks.}

A key assumption of our stop model is that the racial composition of the residential population (as estimated by the U.S.\ Census) is similar to the racial composition of pedestrians officers encounter on the street.
We test how sensitive our inferred thresholds are to this assumption by 
refitting the stop model with various estimates of the fraction of white individuals encountered in each precinct.
Letting $c_{\textrm{white},d}$ denote the original Census estimate,
we varied this number from $c_{\textrm{white},d}/2$ to 
$2c_{\textrm{white},d}$. 
The inferred thresholds remain stable, 
with thresholds for blacks and Hispanics consistently lower than for whites:
thresholds for whites varied from 5.7\% to 6.1\%, thresholds for blacks from 0.8\% to 1.2\%, and thresholds for Hispanics from 1.8\% to 2.1\%. 
This stability is in part due to the 
fact that altering assumptions about the base population 
does not change the observed hit rates, which are substantially higher for whites. We also ran the robustness checks outlined in \citet{simoiu2017}: 
posterior predictive checks,
tests for threshold heterogeneity, and
tests for omitted variable bias (SI Figures 5--7). In all cases, the results confirm our main findings. Standard placebo tests cannot be run in this setting because natural placebos (such as month) eliminate all heterogeneity across groups, breaking model identifiability.

As with the frisk model, HMC inference with discriminant distributions yields better model fit than variational inference with beta distributions. Though variational inference fits more quickly (30 seconds vs. 208 seconds), with variational inference the RMSE of stop rates is four times larger (0.9\% vs. 0.2\%), and the RMSE of hit rates is twice as large (1.3\% vs. 0.8\%). (For comparison, the RMSE of the original beta model for stop rates is 0.2\% and the RMSE of the hit rates is 0.7\%). These performance gaps illustrate the value of full Bayesian inference over approximate methods.

\section{CONCLUSION}

We introduced and analyzed discriminant distributions to accelerate threshold tests for discrimination.
The CDF and conditional means of discriminant distributions reduce to simple expressions that are no more difficult to evaluate than the equivalent statistics for normal distributions. Consequently, using discriminant distributions speeds up inference in the threshold test by more than 75-fold. 
It is now practical to use the threshold test to investigate bias in a wide variety of settings. 
Practitioners can quickly carry out analysis---including computationally expensive robustness checks---on low-cost hardware within minutes.
Our test also scales to previously intractable datasets, such as the national traffic stop database of \citet{pierson_2017}. We also extended the threshold test to domains in which actions are only partially observed, allowing us to assess possible discrimination in an officer's decision to stop a pedestrian. 

Tools for black box Bayesian inference are allowing inference for increasingly complicated models. 
As researchers embrace this complexity, there is opportunity to consider new distributions. 
Historical default distributions---often selected for convenient properties like conjugacy---may not be the best choices when using automatic inference. An early example of this was the Kumaraswamy distribution~\citep{kumaraswamy1980generalized,jones2009kumaraswamy}, developed as an alternative to the beta distribution for its simpler CDF.
Discriminant distributions may also offer computational speedups beyond the threshold test as automatic inference enjoys increasingly widespread use. 

\textbf{Code and acknowledgments: } Code is available at https://github.com/5harad/fasttt. We thank Peng Ding, Avi Feller, Pang Wei Koh, and the reviewers for helpful comments, and the John S. and James L. Knight, Hertz, and NDSEG Foundations.

\balance
\bibliographystyle{plainnat}
\bibliography{references}

\end{document}